\documentclass{llncs}
\usepackage[T1]{fontenc}
\usepackage{graphicx}
\usepackage{amsmath,esvect}
\usepackage{multirow}
\usepackage{amssymb}
\usepackage{microtype}
\usepackage{float}
\usepackage{hyperref} 
\usepackage{color} 
\definecolor{myblue}{RGB}{60,120,216}
\definecolor{revred}{RGB}{190,30,45}

\hypersetup{
  colorlinks=true,
  linkcolor=myblue,
  citecolor=myblue,
  urlcolor=myblue
}
\begin{document}

\title{DADKC: Distribution-Aware Adaptive Differentiated Knowledge Consolidation for Cloth-Hybrid Lifelong Person Re-identification}

\author{Yan Zhang\inst{1} \and
Minghao Long\inst{1} \and
Meiqi Wan\inst{1} \and
Yan Yang\inst{3,4} \and \\
Xin Yuan\inst{1,2,}$^{*}$\and
Wei Wang\inst{1,2} \and
Xiao Wang\inst{1,2}}

\institute{School of Computer Science and Technology, Wuhan University of Science and Technology, Wuhan 430065, China \and
Hubei Province Key Laboratory of Intelligent Information Processing and Real-Time Industrial System, Wuhan University of Science and Technology, Wuhan 430065, China \and
State Key Laboratory of Robotics and Intelligent Systems, Shenyang Institute of Automation, Chinese Academy of Sciences, Shenyang 110016, China \and
China University of Chinese Academy of Sciences, Beijing 100049, China}




\renewcommand{\thefootnote}{}
\footnotetext{$^{*}$ Corresponding author: Xin Yuan (yuanxincherry@gmail.com)}

\maketitle             

\begin{abstract}

Lifelong Person Re-identification (LReID) aims to match the same person using sequentially collected data. In practical long-term deployments, however, the model must handle both domain shifts across cameras and scenes, and irregular clothing changes of the same identity. This gives rise to the Cloth-Hybrid Lifelong Person Re-identification (CH-LReID) task, which demands balancing cloth-relevant and cloth-irrelevant knowledge in streaming data. While existing LReID methods have explored replay-based or distillation-based anti-forgetting strategies, they typically fail to sufficiently model stage-aware domain distributions and cross-stage feature continuity under exemplar-free training. To address this problem, we propose a Distribution-Aware Adaptive Differentiated Knowledge Consolidation (DADKC) framework. The core idea is to enhance the existing Differentiated Knowledge Consolidation (DKC) paradigm with adaptive distribution modeling. To this end, three lightweight modules are designed: (1) an Anchor Pool Adapter (APA) to retrieve learnable anchor cues for refining identity features; (2) a Phase Path Mixer (PPM) to perform stage-aware expert routing for integrating domain-specific distribution paths; and (3) a Stage Bridge Alignment (SBA) module to regularize the continuity of base, anchor, and path representations between adjacent stages. 
Extensive experiments on the CH-LReID and LReID benchmarks show that DADKC gains 4.1\%/4.5\% mAP/R1 on Order-1 and 3.6\%/3.4\% mAP/R1 on Order-2 over the DKC baseline, and lowers weighted forgetting from 2.4\%/1.7\% to 1.8\%/1.4\%, while storing neither historical images nor per-image features. The code is available at \url{https://github.com/fire667/DADKC}.


\end{abstract}      

\section{Introduction}

Lifelong person re-identification (LReID) aims to match the same person by continuously learning from sequentially collected data~\cite{jiang2025multi,pu2021aka,xu2024dkp,xu2022rank}. Similar to most lifelong learning methods~\cite{xu2024lstkc,wu2021gwf}, LReID typically suffers from the catastrophic forgetting problem~\cite{delange2021survey,masana2022classincremental}. To this end, replay-based LReID methods~\cite{pu2021aka,yu2023krc} preserve old knowledge by rehearsing historical data, while non-exemplar LReID methods~\cite{li2024pgakc,xu2024dask} employ knowledge distillation~\cite{sun2022patchkd} to avoid the involvement of historical data due to privacy issues~\cite{xu2024dkp,xu2022rank,yuan2026calibration}. These methods have achieved promising progress in cloth-consistent scenarios, where clothing style remains a stable and discriminative cue.

\begin{figure*}[!b]
    \centering
    \includegraphics[width=0.8\textwidth]{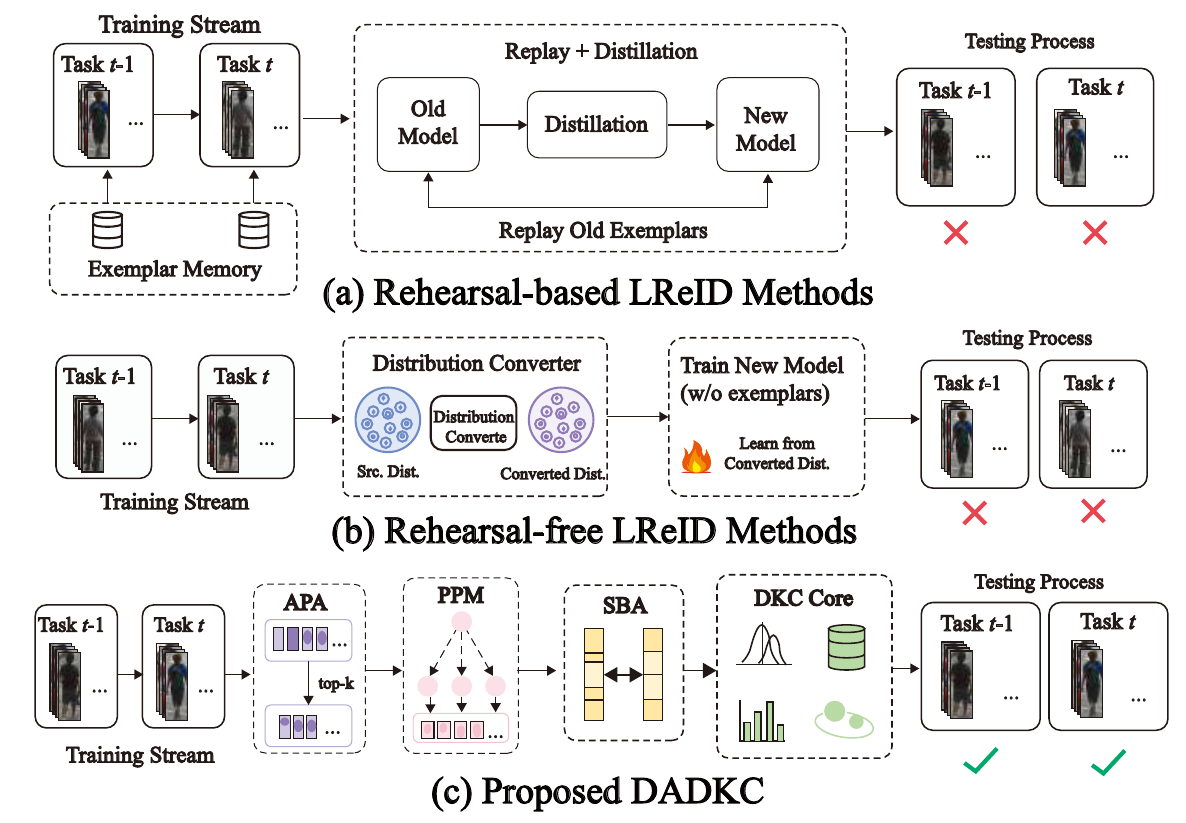}
    \caption{Paradigm comparison of LReID methods: rehearsal-based methods replay exemplars, rehearsal-free methods avoid raw samples but may lose cross-stage continuity, and DADKC performs distribution-aware differentiated consolidation.}
    \label{fig:motivation}
    \vspace{-2mm}
\end{figure*}

However, considering the long-term nature of LReID, human clothing often changes irregularly across days or seasons~\cite{li2025event,qian2020ltcc,xu2023deepchange,yuan2023searching,tian2026structural}, which hinders the model from relying on unified discriminative information. As shown in Fig.~\ref{fig:motivation}(a), the appearance variety of the same person due to the cloth-changing issue requires the acquisition and anti-forgetting of identity discriminative knowledge from both cloth-consistent and cloth-changing data, termed the cloth-hybrid scenario. Specifically, cloth-relevant information (\textit{e.g.}, clothing style) is often discriminative in cloth-consistent data, but it is not always effective or even harmful in cloth-changing ones. Existing methods~\cite{wu2021gwf,xu2024lstkc,yu2023krc} fail to balance such significantly differentiated knowledge since they typically exacerbate the catastrophic knowledge conflict challenge between cloth-relevant and cloth-irrelevant cues. Although the recent DKC~\cite{cui2025dkc} introduces differentiated knowledge consolidation to address this conflict, its distribution modeling capacity is still limited: it does not explicitly handle stage-aware domain shifts (\textit{e.g.}, illumination, camera style) and cross-stage feature continuity under exemplar-free training. As shown in Fig.~\ref{fig:motivation}(b), without proper distribution-aware adaptation, the model fails to preserve both historical knowledge and new domain-specific representations. 
DADKC therefore extends DKC rather than proposing yet another continual ReID paradigm, and its distinction is not a generic memory, router, or alignment loss added in isolation. The three components are coupled inside the differentiated-consolidation pipeline: APA attaches retrieved anchors to feature and uncertainty refinement, PPM conditions expert routing on those anchors, and SBA constrains the continuity of the resulting mean, anchor, and path representations.

Motivated by these observations, we propose DADKC for Cloth-Hybrid Lifelong Person Re-identification (CH-LReID), as shown in Fig.~\ref{fig:motivation}(c). Built on the ResNet-based DKC framework, it couples three lightweight modules: APA retrieves learnable anchors to refine identity features without storing historical exemplars; PPM integrates domain-specific paths through stage-aware expert routing; and SBA aligns representations across adjacent stages to reduce drift. The main contributions are as follows:

\begin{itemize}
\item A Distribution-Aware Adaptive Differentiated Knowledge Consolidation (DADKC) framework is proposed to solve the Cloth-Hybrid Lifelong Person Re-identification task, which requires matching the same person using streaming cloth-consistent and cloth-changing data while handling domain shifts and clothing variations.

\item An Anchor Pool Adapter (APA) and a Phase Path Mixer (PPM) are proposed to enhance the DKC representation with learnable anchor guidance and stage-aware expert path integration, improving distribution modeling without storing old exemplars.

\item A Stage Bridge Alignment (SBA) module is designed to regularize cross-stage continuity over base, anchor, and path representations, complementing prototype-based and differentiated knowledge consolidation losses.

\item Extensive experiments on CH-LReID and LReID benchmarks show that DADKC improves average retrieval performance over the DKC baseline, with reduced aggregate forgetting on the evaluated CH-LReID streams.
\end{itemize}

\section{Related Work}

\subsection{Lifelong Person Re-identification}

Lifelong person re-identification (LReID) trains ReID models on sequential domains while retaining historical performance, with catastrophic forgetting as the central challenge~\cite{delange2021survey,masana2022classincremental}. Replay-based methods store exemplars for rehearsal: AKA~\cite{pu2021aka} accumulates knowledge with an adaptive graph, GwF~\cite{wu2021gwf} preserves classification, distribution and representation coherence, KRC~\cite{yu2023krc} refreshes representative samples, PatchKD~\cite{sun2022patchkd} distills local patches, and CCR~\cite{cui2024ccr} maintains compatible embeddings for re-indexing-free updates. These methods are effective but constrained by privacy and storage.

Exemplar-free methods avoid raw-image memory by consolidating parameters, prototypes, or domain styles. PGAKC~\cite{li2024pgakc} estimates old-knowledge importance under prompt guidance, DKP~\cite{xu2024dkp} rehearses generated identity prototypes, DASK~\cite{xu2024dask} stores adaptive style kernels, ASK~\cite{qian2024ask} selects adapters by distribution similarity, and DAFC~\cite{liu2025dafc} integrates domain-specific and shared representations. However, they are mainly designed for cloth-consistent streams and do not explicitly handle the conflict between cloth-relevant and cloth-irrelevant cues.

\subsection{Cloth-Hybrid Lifelong Person Re-identification}

Cloth-changing ReID (CC-ReID) matches identities across outfits by reducing clothing bias and exploiting stable cues such as body shape or gait~\cite{qian2020ltcc,wang2020re,xu2023deepchange,yuan2025identity}. LTCC~\cite{qian2020ltcc} introduces a long-term RGB-contour benchmark, DeepChange~\cite{xu2023deepchange} expands temporal and scale diversity, and DCR-ReID~\cite{cui2023dcr} reconstructs identity features from cloth-insensitive body components. These static-dataset methods do not address representation drift in sequential learning. CH-LReID is harder because clothing cues are useful in cloth-consistent stages but unreliable in cloth-changing stages. DKC~\cite{cui2025dkc} defines this task and uses clustering-based differentiated transfer and latent knowledge consolidation, while DADKC further adds lightweight distribution-aware adaptation and cross-stage continuity modeling.

\subsection{Distribution-Aware Continual Learning}

Distribution-aware continual learning models task-specific feature distributions without historical replay~\cite{delange2021survey,masana2022classincremental}. Prompt methods~\cite{smith2023codaprompt,wang2022sprompt,wang2022dualprompt,wang2022l2p} adapt frozen backbones with learnable prompts, while mixture-of-experts models~\cite{riquelme2021moe,yu2024moe} expand capacity through dynamic routing. In exemplar-free LReID, DKP~\cite{xu2024dkp}, DASK~\cite{xu2024dask}, and DAFC~\cite{liu2025dafc} model identity prototypes, domain styles, and global distributions, respectively. Yet these approaches usually treat ReID features as a unified distribution. DADKC instead implements ResNet-compatible APA, PPM, and SBA modules on top of differentiated knowledge consolidation~\cite{cui2025dkc}, so stage-aware domain shifts and cross-stage continuity are modeled in cloth-hybrid streams. 
The difference from these methods lies in what they require: DASK rehearses style-transferred distributions, ASK selects adapters separately for each test domain, and DAFC adds a text/CLIP branch, whereas DADKC needs no generated samples, no text supervision, and no test-time domain selection.

\section{Method}

\subsection{Preliminary}

In cloth-hybrid lifelong person re-identification, training data arrive as stages $\mathcal{D}=\{\mathcal{D}^{1},\mathcal{D}^{2},\ldots,\mathcal{D}^{T}\}$. At stage $t$, only $\mathcal{D}^{t}=\{(x_i^{t},y_i^{t},d_i^{t})\}_{i=1}^{N_t}$ is available, where $x_i^{t}$, $y_i^{t}$, and $d_i^{t}$ are the image, identity label, and domain index. Each benchmark stage contains one dataset/domain, so $d_i^t$ is constant within a stage. We distinguish stage from domain: the former controls expert availability, and the latter is only an optional routing cue. Historical images are inaccessible, and inference requires neither identity nor domain labels. The goal is to adapt $\phi^t$ to the current stage while preserving earlier discriminative knowledge.

Compared with standard LReID, CH-LReID mixes cloth-consistent and cloth-changing streams. Thus, the model must preserve cloth-relevant cues in cloth-consistent stages while learning cloth-irrelevant cues robust to clothing variation. DADKC handles the resulting forgetting and knowledge conflict by adding distribution-aware adaptive modeling to differentiated knowledge consolidation.

\subsection{Overall Framework}

\begin{figure*}[t]
    \centering
    \includegraphics[width=1.0\textwidth]{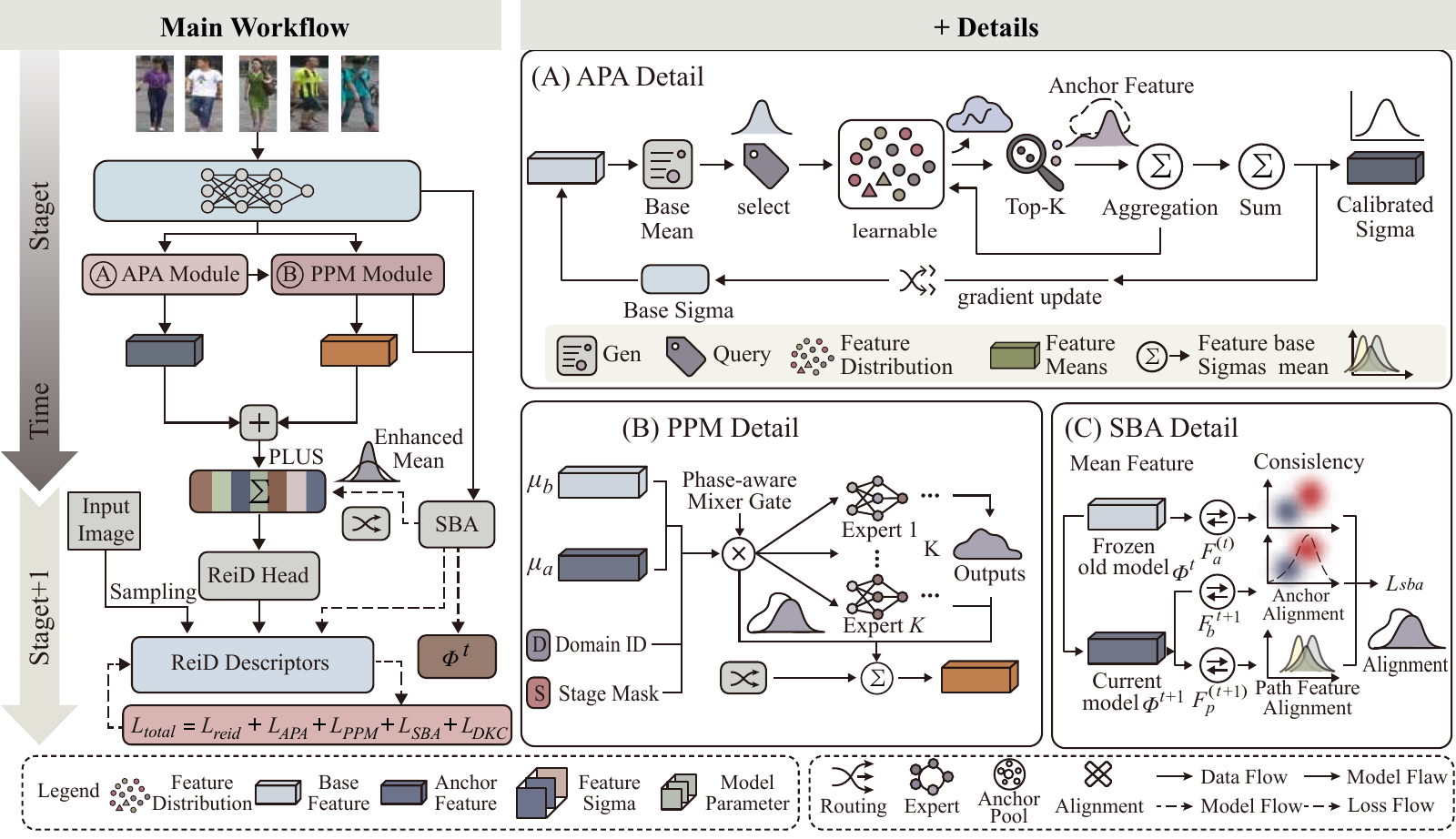}
    \caption{Overall framework of DADKC. A DKC-style backbone extracts base and uncertainty representations; APA retrieves anchor cues, PPM routes through stage-aware experts, and SBA aligns current and previous-stage base, anchor, and path features. The final model uses accumulated experts and an unknown-domain embedding at test time.}
    \label{fig:framework}
\end{figure*}

As shown in Fig.~\ref{fig:framework}, DADKC follows the DKC-style continual ReID pipeline and introduces three lightweight modules: Anchor Pool Adapter (APA), Phase Path Mixer (PPM), and Stage Bridge Alignment (SBA). The backbone extracts base and uncertainty representations; APA retrieves anchor cues; PPM routes through stage-aware experts; and SBA aligns current and historical base, anchor, and path features. Historical experts are frozen after their stages, while current experts, router rows, projection heads, and the anchor pool remain trainable. The fused representation is optimized with standard ReID losses, APA/PPM/SBA regularization, anti-forgetting loss, and the DKC consistency branch~\cite{cui2025dkc}.

\subsection{Distribution-Aware Representation}

Following DKC~\cite{cui2025dkc}, DADKC uses a ResNet-based backbone to produce a feature map, a normalized base feature, and a diagonal uncertainty vector:
\begin{equation}
\begin{aligned}
    \mathbf{z}_i^{t}&=f_{\theta}^{t}(x_i^{t}), &
    \mathbf{b}_i^{t}&=\operatorname{Norm}(g_{\mu}(\mathbf{z}_i^{t})), &
    \boldsymbol{\sigma}_i^{t}&=\operatorname{ReLU}(g_{\sigma}(\mathbf{z}_i^{t}))+\epsilon .
\end{aligned}
\end{equation}
Here, $\epsilon$ ensures numerical stability. The estimated uncertainty is later used to sample features for classification and triplet learning.

APA maintains learnable anchor keys $\mathbf{K}=\{\mathbf{k}_m\}_{m=1}^{M}$ and values $\mathbf{A}=\{\mathbf{a}_m\}_{m=1}^{M}$. Given $\mathbf{b}_i^t$, it forms a normalized query $\mathbf{q}_i^t=\operatorname{Norm}(W_q(\mathbf{b}_i^t))$, computes softmax similarities $s_{i,m}^t$ to the normalized anchor keys, and aggregates the re-normalized top-$K$ anchors as $\mathbf{a}_i^t=\sum_{m\in \mathcal{T}_K(i)}\bar{s}_{i,m}^t\mathbf{a}_m$. To make anchor cues identity-aware and uncertainty-aware, APA uses
\begin{equation}
    \mathcal{L}_{apa}=1-\frac{1}{B}\sum_{i=1}^{B}
    \operatorname{cos}(\mathbf{a}_i^{t},\mathbf{b}_i^{t}),\quad
    \hat{\boldsymbol{\sigma}}_i^{t}
    =\boldsymbol{\sigma}_i^{t}\odot(1+\eta G_a(\mathbf{a}_i^{t})).
\end{equation}
Here, $\mathcal{T}_K(i)$ is the selected anchor set, $\bar{s}_{i,m}^{t}$ is its re-normalized score, $G_a(\cdot)$ is a sigmoid gate, $B$ is the mini-batch size, and $\eta$ controls the calibration strength.

\subsection{Phase Path Mixer and Sampling}

PPM appends $E$ bottleneck experts at each stage and retains earlier frozen experts. For each sample, the router takes $\mathbf{u}_i^t=[\mathbf{b}_i^t;\mathbf{a}_i^t;\mathbf{e}_{d_i}^t]$ as input; the observed domain embedding is replaced by a shared unknown-domain embedding with probability $p_u$ so routing remains label-free at test time. 
Stage information only fixes which experts are active during training; at inference, all accumulated experts are available and routing requires neither the domain label nor the stage label of the test image.
With the stage mask $\mathbf{m}^t$, the active weights are $\boldsymbol{\pi}_i^t=\operatorname{Softmax}(G_p(\mathbf{u}_i^t)+\mathbf{m}^t)$, and the global path feature is $\mathbf{p}_i^t=\sum_{r=1}^{R_t}\pi_{i,r}^tE_r(\mathbf{b}_i^t)$. We regularize routing by
\begin{equation}
    \mathcal{L}_{ppm}=1-\frac{1}{B}\sum_{i=1}^{B}\operatorname{cos}(\mathbf{p}_{i,s}^{t},\hat{\boldsymbol{\mu}}_i^{t})
    +\lambda_{e}\mathcal{H}(\boldsymbol{\pi}_i^{t}).
\end{equation}
The global path $\mathbf{p}_i^t$ integrates historical and current-stage distribution knowledge, while the stage-specific path $\mathbf{p}_{i,s}^t$ stabilizes the newly learned distribution.

The final mean representation is $\hat{\boldsymbol{\mu}}_i^t=\operatorname{Norm}(\mathbf{b}_i^t+\alpha\mathbf{a}_i^t+\beta\mathbf{p}_i^t)$, and Gaussian sampling follows the uncertainty-aware DKC protocol as $\tilde{\mathbf{h}}_{i,k}^t=\hat{\boldsymbol{\mu}}_i^t+\boldsymbol{\epsilon}_{k}\odot\hat{\boldsymbol{\sigma}}_i^t$, where $\boldsymbol{\epsilon}_{k}\sim\mathcal{N}(\mathbf{0},\mathbf{I})$. The enhanced mean and sampled features are jointly sent to the classifier and triplet loss, so anchor and path cues participate in the main ReID representation.

\subsection{Stage Bridge Alignment and Objective}

Before optimizing stage $t>1$, DADKC snapshots the converged stage-$(t-1)$ model as a frozen teacher. SBA feeds current images into teacher and student, stops gradients through teacher outputs, uses the shared unknown-domain embedding for both models, and sets $\mathcal{L}_{sba}=0$ at $t=1$. It aligns base mean, anchor, and path representations:
\begin{equation}
\begin{aligned}
    \mathcal{L}_{sba}&=\frac{1}{3}(\mathcal{L}_{mean}+\mathcal{L}_{anchor}+\mathcal{L}_{path}),\\
    \mathcal{L}_{mean}&=1-\frac{1}{B}\sum_{i=1}^{B}
    \operatorname{cos}(\hat{\boldsymbol{\mu}}_i^{t},\boldsymbol{\mu}_i^{t-1}),\\
    \mathcal{L}_{anchor}&=1-\frac{1}{B}\sum_{i=1}^{B}
    \operatorname{cos}(\mathbf{a}_i^{t},\mathbf{a}_i^{t-1}),\quad
    \mathcal{L}_{path}=1-\frac{1}{B}\sum_{i=1}^{B}
    \operatorname{cos}(\mathbf{p}_i^{t},\mathbf{p}_i^{t-1}).
\end{aligned}
\end{equation}
Thus, SBA complements DKC's relation-level consistency by constraining the newly introduced adaptive branches.

DADKC also preserves the DKC consistency branch~\cite{cui2025dkc}: historical prototype memory stores class-level mean and variance statistics, sampled prototypes define $\mathcal{L}_{af}$, relation matrices from current and previous models are reconstructed and aligned, and DBSCAN-updated fine-grained pseudo labels provide auxiliary cloth-hybrid structure. The optimization objective is
\begin{equation}
\begin{aligned}
    \mathcal{L}_{dkc}&=\mathcal{L}_{div1}+\mathcal{L}_{div2}+\mathcal{L}_{div3}+\mathcal{L}_{div4},\quad
    \mathcal{L}_{reid}=\mathcal{L}_{ce}+\mathcal{L}_{tri},\\
    \mathcal{L}_{total}
    &=\mathcal{L}_{reid}
    +\lambda_{apa}\mathcal{L}_{apa}
    +\lambda_{ppm}\mathcal{L}_{ppm}
    +\lambda_{sba}\mathcal{L}_{sba}
    +\lambda_{af}\mathcal{L}_{af}
    +\mathcal{L}_{dkc}.
\end{aligned}
\end{equation}
At test time, the final stage mask activates all accumulated experts, the router uses the shared unknown-domain embedding, and the deterministic enhanced mean $\hat{\boldsymbol{\mu}}$ is used as the descriptor. No identity label, domain label, historical image, clustering information, teacher model, or stochastic sample is required. 
After $T$ stages, the persistent state is the anchor pool ($Md$ values), the expert bank ($T P_e$ parameters), the routers ($3d\sum_{t}R_t$ parameters), and the class statistics ($2d\sum_{t}N_t$ values). No image or per-image feature is stored. For the default configuration, the five-stage state is about 53 MB in FP32, dominated by class statistics; the model itself is 31.2M parameters.

\section{Experiments}

\subsection{Datasets and Evaluation Metrics}

\textbf{Datasets.}
Following the DKC cloth-hybrid protocol~\cite{cui2025dkc}, we evaluate on cloth-consistent datasets (Market-1501, CUHK-SYSU, DukeMTMC-ReID, MSMT17-V2, CUHK03) and cloth-changing datasets (LTCC~\cite{qian2020ltcc}, PRCC). The CH-LReID orders are \textit{Order-1}: Market-1501 $\rightarrow$ LTCC $\rightarrow$ PRCC $\rightarrow$ MSMT17-V2 $\rightarrow$ CUHK03, and \textit{Order-2}: MSMT17-V2 $\rightarrow$ PRCC $\rightarrow$ Market-1501 $\rightarrow$ CUHK03 $\rightarrow$ LTCC. Following DKC~\cite{cui2025dkc} and previous LReID works~\cite{pu2021aka,xu2024dkp}, we also evaluate standard LReID with \textit{Order-3}: Market-1501 $\rightarrow$ CUHK-SYSU $\rightarrow$ DukeMTMC-ReID $\rightarrow$ MSMT17-V2 $\rightarrow$ CUHK03, and \textit{Order-4}: DukeMTMC-ReID $\rightarrow$ MSMT17-V2 $\rightarrow$ Market-1501 $\rightarrow$ CUHK-SYSU $\rightarrow$ CUHK03.

\textbf{Evaluation Metrics.}
We report mean Average Precision (mAP) and Rank-1 accuracy (R1) on each dataset. For lifelong learning evaluation, we further report the average mAP/R1 over all evaluated domains. Following DKC \cite{cui2025dkc}, we also use Average Forgetting (AF) and Backward Transfer (BWT) to measure the sequence-level retention behavior. Let $\mathcal{E}$ denote the seven evaluation protocols, including five same-clothes and two cross-clothes evaluations, let $\tau(e)$ be the stage at which evaluation protocol $e$ is introduced, and let $r_{t,e}$ be the performance on $e$ after training stage $t$. We compute
\begin{equation}
\mathrm{AF}=\frac{1}{|\mathcal{E}|}\sum_{e\in\mathcal{E}}
\left(\max_{\tau(e)\leq l\leq T} r_{l,e}-r_{T,e}\right), \quad
\mathrm{BWT}=\frac{1}{|\mathcal{E}|}\sum_{e\in\mathcal{E}}
\left(r_{T,e}-r_{\tau(e),e}\right).
\end{equation}
Lower AF indicates stronger anti-forgetting ability, while larger BWT indicates better retention or positive backward transfer. Same-clothes and cross-clothes scores are averaged separately, and the overall score uses the $5:2$ weighting, e.g., $\mathrm{AF}_{all}=(5\mathrm{AF}_{same}+2\mathrm{AF}_{cross})/7$.

\subsection{Implementation Details}

DADKC uses a ResNet-50 backbone initialized from ImageNet-pretrained weights. Images are resized to $256\times128$ and augmented with random flip, random crop, and random erasing. The backbone and classifier are trained with cross-entropy loss and triplet loss. Following the DKC setting \cite{cui2025dkc}, the first stage is trained for 80 epochs, and the remaining stages are trained for 60 epochs each. We use SGD with momentum 0.9, weight decay $5\times10^{-4}$, initial learning rate 0.008, warm-up for 10 epochs, and 0.1 decay at epochs 40 and 70 in the first stage and epochs 30 and 50 in later stages. Each batch contains 32 identities and 4 images per identity, giving a batch size of 128. For each dataset, 500 identities are randomly sampled during training when enough identities are available; otherwise, all identities are used. DBSCAN uses $\epsilon=0.6$ and min\_samples $=4$, and fine-grained pseudo labels are updated every 10 epochs following the same protocol.

For the DADKC-specific modules, the Anchor Pool Adapter (APA), Phase Path Mixer (PPM), and Stage Bridge Alignment (SBA) are optimized jointly with the DKC consistency branch. Unless otherwise specified, we set the anchor pool size $M$ to 128, the number of retrieved anchors $K$ to 8, the number of experts per stage $E$ to 5, the number of Gaussian samples $S$ to 4, $\alpha=\beta=0.1$, $\eta=0.1$, and $\lambda_e=0.01$. The loss weights $\lambda_{apa}$, $\lambda_{ppm}$, $\lambda_{sba}$, $\lambda_{af}$, and $\lambda_{dkc}$ are set to 0.1, 0.1, 0.01, 1, and 1, respectively. Newly run DADKC and DKC-controlled results use seeds $\{1,11,21,31,41\}$ and are reported as five-seed means. All implementation details follow the released DKC protocol unless otherwise stated.

\begin{table*}[t]
\centering
\caption{Order-1 performance. DKC-reported values are from DKC \cite{cui2025dkc}; DKC-controlled and DADKC are common-code five-seed means.}
\label{tab:order1}
\resizebox{\textwidth}{!}{
\begin{tabular}{l|cc|cc|cc|cc|cc|cc|cc|cc|cc|cc}
\hline
\multirow{3}{*}{Method} &
\multicolumn{12}{c|}{Cloth-Consistent} &
\multicolumn{6}{c|}{Cloth-Changing} &
\multicolumn{2}{c}{Average} \\
\cline{2-21}
& \multicolumn{2}{c|}{Market} & \multicolumn{2}{c|}{LTCC} & \multicolumn{2}{c|}{PRCC} & \multicolumn{2}{c|}{MSMT17} & \multicolumn{2}{c|}{CUHK03}
& \multicolumn{2}{c|}{Avg.} & \multicolumn{2}{c|}{LTCC} & \multicolumn{2}{c|}{PRCC} & \multicolumn{2}{c|}{Avg.} & mAP & R1 \\
\cline{2-21}
& mAP & R1 & mAP & R1 & mAP & R1 & mAP & R1 & mAP & R1 & mAP & R1 & mAP & R1 & mAP & R1 & mAP & R1 & & \\
\hline
Joint & 64.1 & 82.5 & 42.6 & 62.1 & 94.6 & 98.7 & 18.4 & 40.8 & 44.4 & 46.4 & 52.8 & 66.1 & 10.1 & 23.0 & 32.7 & 33.8 & 21.4 & 28.4 & 43.8 & 55.3 \\
SFT & 28.5 & 52.0 & 28.5 & 49.3 & 92.5 & 97.3 & 7.0 & 19.6 & 44.0 & 45.6 & 40.1 & 52.8 & 6.9 & 15.3 & 21.8 & 21.2 & 14.4 & 18.3 & 32.7 & 42.9 \\
AKA (CVPR 2021)~\cite{pu2021aka} & 48.0 & 69.5 & 25.4 & 45.1 & 88.1 & 93.3 & 4.2 & 12.0 & 31.2 & 31.2 & 39.4 & 50.2 & 6.5 & 12.8 & 26.5 & 26.7 & 16.5 & 19.8 & 32.8 & 41.5 \\
PatchKD (ACM MM 2022)~\cite{sun2022patchkd} & \textbf{68.0} & \textbf{85.5} & 30.8 & 54.7 & 93.5 & 96.5 & 5.7 & 15.6 & 33.2 & 32.9 & 46.2 & 57.0 & 7.2 & 17.9 & 26.1 & 26.0 & 16.7 & 22.0 & 37.8 & 47.0 \\
LSTKC (AAAI 2024)~\cite{xu2024lstkc} & 39.9 & 63.4 & 39.6 & 65.4 & 95.9 & 98.9 & 11.5 & 29.2 & \textbf{48.1} & \textbf{50.1} & 47.0 & 61.4 & 8.3 & 19.4 & 24.0 & 22.9 & 16.2 & 21.2 & 38.2 & 49.9 \\
USP (TCSVT 2024)~\cite{yan2024unified} & \underline{66.3} & 74.9 & 37.2 & 49.6 & 88.8 & 92.6 & 6.6 & 14.6 & 38.7 & 33.9 & 47.5 & 53.1 & 7.8 & 17.3 & 25.7 & 23.6 & 16.8 & 20.5 & 38.7 & 43.8 \\
DKP (CVPR 2024)~\cite{xu2024dkp} & 51.3 & 73.0 & 44.2 & 68.6 & 98.4 & 99.4 & 13.5 & 31.6 & 39.5 & 40.6 & 49.4 & 62.6 & 9.3 & 21.4 & 35.5 & 34.8 & 22.4 & 28.1 & 41.7 & 52.8 \\
DKC-reported~\cite{cui2025dkc} & 57.9 & 78.5 & \underline{49.2} & \underline{72.4} & \underline{98.6} & \underline{99.5} & \underline{16.2} & \underline{36.7} & 40.7 & 42.4 & \underline{52.5} & \underline{65.9} & \underline{10.1} & \underline{25.5} & \underline{36.9} & 35.5 & \underline{23.5} & \underline{30.5} & \underline{44.2} & \underline{55.8} \\
DKC-controlled & 54.4 & 76.1 & 44.2 & 70.0 & 98.2 & 99.4 & 14.2 & 33.4 & 34.1 & 34.2 & 49.0 & 62.6 & 8.9 & 20.4 & \underline{36.9} & \underline{37.9} & 22.9 & 29.2 & 41.6 & 53.1 \\
DADKC (Ours) & 59.3 & \underline{79.6} & \textbf{51.4} & \textbf{74.1} & \textbf{98.9} & \textbf{99.6} & \textbf{17.8} & \textbf{38.9} & \underline{42.6} & \underline{44.7} & \textbf{54.0} & \textbf{67.4} & \textbf{11.7} & \textbf{27.8} & \textbf{38.2} & \textbf{38.5} & \textbf{25.0} & \textbf{33.2} & \textbf{45.7} & \textbf{57.6} \\
\hline
\end{tabular}}
\end{table*}

\begin{table*}[t]
\centering
\caption{Order-2 performance. DKC-reported values are from DKC \cite{cui2025dkc}; DKC-controlled and DADKC are common-code five-seed means.}
\label{tab:order2}
\resizebox{\textwidth}{!}{
\begin{tabular}{l|cc|cc|cc|cc|cc|cc|cc|cc|cc|cc}
\hline
\multirow{3}{*}{Method} &
\multicolumn{12}{c|}{Cloth-Consistent} &
\multicolumn{6}{c|}{Cloth-Changing} &
\multicolumn{2}{c}{Average} \\
\cline{2-21}
& \multicolumn{2}{c|}{MSMT17} & \multicolumn{2}{c|}{PRCC} & \multicolumn{2}{c|}{Market} & \multicolumn{2}{c|}{CUHK03} & \multicolumn{2}{c|}{LTCC}
& \multicolumn{2}{c|}{Avg.} & \multicolumn{2}{c|}{PRCC} & \multicolumn{2}{c|}{LTCC} & \multicolumn{2}{c|}{Avg.} & mAP & R1 \\
\cline{2-21}
& mAP & R1 & mAP & R1 & mAP & R1 & mAP & R1 & mAP & R1 & mAP & R1 & mAP & R1 & mAP & R1 & mAP & R1 & & \\
\hline
Joint & 18.4 & 40.8 & 94.6 & 98.7 & 64.1 & 82.5 & 44.4 & 46.4 & 42.6 & 62.1 & 52.8 & 66.1 & 32.7 & 33.8 & 10.1 & 23.0 & 21.4 & 28.4 & 43.8 & 55.3 \\
SFT & 2.4 & 9.1 & 84.5 & 95.9 & 16.9 & 37.2 & 8.2 & 8.8 & 37.9 & 57.6 & 30.0 & 41.7 & 21.9 & 24.3 & 9.1 & 20.9 & 15.5 & 22.6 & 25.8 & 36.3 \\
AKA (CVPR 2021)~\cite{pu2021aka} & 15.0 & 32.4 & 93.0 & 97.0 & 26.6 & 51.0 & 13.4 & 11.6 & 43.4 & 64.0 & 38.3 & 51.2 & 28.2 & 27.7 & 8.4 & 15.8 & 18.3 & 21.8 & 32.6 & 42.8 \\
PatchKD (ACM MM 2022)~\cite{sun2022patchkd} & \underline{22.6} & \textbf{47.9} & 95.0 & 98.5 & 38.2 & 64.2 & 21.8 & 23.1 & 45.8 & 70.0 & 44.7 & 60.7 & 28.1 & 27.6 & 9.3 & 21.7 & 18.7 & 24.7 & 37.3 & 50.4 \\
LSTKC (AAAI 2024)~\cite{xu2024lstkc} & 6.9 & 20.4 & 92.8 & 97.0 & 39.8 & 64.3 & 25.2 & 26.0 & 51.1 & 65.5 & 43.2 & 54.6 & 27.3 & 27.0 & 10.9 & 22.4 & 19.1 & 24.7 & 36.3 & 46.1 \\
USP (TCSVT 2024)~\cite{yan2024unified} & \textbf{22.7} & 37.2 & 92.5 & 95.9 & 37.8 & 53.5 & 17.3 & 14.6 & \textbf{63.5} & 72.4 & 46.8 & 54.7 & 28.7 & 29.4 & 11.9 & 26.0 & 20.3 & 27.7 & 39.2 & 47.0 \\
DKP (CVPR 2024)~\cite{xu2024dkp} & 11.8 & 29.0 & 96.7 & 98.8 & 46.0 & 69.8 & 29.9 & 30.9 & 55.8 & 75.8 & 48.0 & 60.9 & 34.1 & 34.6 & 11.6 & 24.2 & 22.9 & 29.4 & 40.8 & 51.9 \\
DKC-reported~\cite{cui2025dkc} & 15.4 & 35.3 & 98.1 & 99.0 & \underline{51.8} & \underline{72.7} & \underline{31.9} & \underline{32.1} & 57.2 & \underline{77.9} & \underline{50.9} & \underline{63.4} & 38.2 & 37.9 & \underline{13.0} & \underline{27.0} & \underline{25.6} & \underline{32.5} & \underline{43.7} & \underline{54.6} \\
DKC-controlled & 15.3 & 36.3 & \underline{98.3} & \textbf{99.9} & 48.2 & 70.9 & 28.3 & 27.9 & 51.3 & 71.0 & 48.3 & 61.2 & \underline{38.3} & \textbf{40.1} & 11.6 & 24.5 & 24.9 & 32.3 & 41.6 & 52.9 \\
DADKC (Ours) & 17.1 & \underline{37.4} & \textbf{98.6} & \underline{99.4} & \textbf{53.5} & \textbf{74.2} & \textbf{33.8} & \textbf{34.6} & \underline{59.0} & \textbf{79.2} & \textbf{52.4} & \textbf{65.0} & \textbf{39.7} & \underline{40.0} & \textbf{14.5} & \textbf{29.1} & \textbf{27.1} & \textbf{34.6} & \textbf{45.2} & \textbf{56.3} \\
\hline
\end{tabular}}
\end{table*}

\subsection{Comparison with State-of-the-art Methods}

\textbf{Compared Methods.}
We compare DADKC with AKA \cite{pu2021aka}, PatchKD \cite{sun2022patchkd}, LSTKC \cite{xu2024lstkc}, USP \cite{yan2024unified}, DKP \cite{xu2024dkp}, and DKC \cite{cui2025dkc}. Since DADKC is built on the DKC protocol, Tables~\ref{tab:order1} and~\ref{tab:order2} report both the published DKC result and a common-code DKC control. DKC-controlled and DADKC use the same five random seeds. In all tables, \textbf{bold} and \underline{underlining} mark the best and second-best results, respectively; Joint and SFT are excluded for Orders 1 and 2.

As shown in Tables~\ref{tab:order1} and~\ref{tab:order2}, DKC provides a strong baseline for the CH-LReID task by balancing cloth-relevant and cloth-irrelevant knowledge across sequential stages. DADKC obtains 45.7\%/57.6\% mAP/R1 on Order-1 and 45.2\%/56.3\% mAP/R1 on Order-2, exceeding the published DKC results by 1.5\%/1.8\% and 1.5\%/1.7\% mAP/R1, respectively. Across five seeds, the aggregate standard deviations are at most 0.4\%: Order-1 is $45.7\pm0.3\%/57.6\pm0.4\%$ mAP/R1 and Order-2 is $45.2\pm0.4\%/56.3\pm0.3\%$ mAP/R1. Therefore, the comparison focuses on whether these distribution-aware components can improve the average mAP/R1 and reduce performance degradation in both cloth-consistent and cloth-changing domains.
Table~\ref{tab:order34} reports the LReID setting composed only of cloth-consistent datasets. This evaluation is complementary to the CH-LReID setting: it examines whether DADKC preserves the lifelong ReID ability inherited from DKC while introducing additional distribution-aware components for cloth-hybrid streams.

\textbf{Anti-forgetting Analysis.}
To further evaluate anti-forgetting ability, we report the AF score on training \textit{Order-1}. Lower values indicate that the method better preserves historical knowledge after learning later domains. Table~\ref{tab:af} reports same-clothes, cross-clothes, and weighted seven-protocol aggregate results.

\begin{table}[!htbp]
\centering
\caption{Traditional cloth-consistent LReID results. Baselines are from DKC \cite{cui2025dkc}; DKC-controlled and DADKC are five-seed means.}
\label{tab:order34}
\scriptsize
\setlength{\tabcolsep}{4.0pt}
\renewcommand{\arraystretch}{1.12}
\begin{tabular}{l|cc|cc|cc}
\hline
\multirow{2}{*}{Method} & \multicolumn{2}{c|}{Order-3} & \multicolumn{2}{c|}{Order-4} & \multicolumn{2}{c}{Average} \\
\cline{2-7}
& mAP & R1 & mAP & R1 & mAP & R1 \\
\hline
CRL (WACV 2021)~\cite{zhao2021continual} & 40.5 & 50.8 & 37.6 & 49.2 & 39.1 & 50.0 \\
AKA (CVPR 2021)~\cite{pu2021aka} & 40.8 & 50.8 & 38.6 & 49.4 & 39.7 & 50.1 \\
PatchKD (ACM MM 2022)~\cite{sun2022patchkd} & 43.7 & 53.7 & 43.2 & 54.1 & 43.5 & 53.9 \\
LSTKC (AAAI 2024)~\cite{xu2024lstkc} & 50.0 & 63.1 & 49.6 & 62.1 & 49.8 & 62.6 \\
DKP (CVPR 2024)~\cite{xu2024dkp} & \underline{51.8} & \underline{64.1} & 51.3 & 63.2 & \underline{51.6} & 63.7 \\
DKC-controlled & 51.6 & \underline{64.1} & \underline{51.6} & \underline{63.5} & \underline{51.6} & \underline{63.8} \\
DADKC (Ours) & \textbf{53.2} & \textbf{65.5} & \textbf{53.0} & \textbf{65.2} & \textbf{53.1} & \textbf{65.4} \\
\hline
\end{tabular}
\end{table}

\begin{table}[!htbp]
\centering
\caption{Order-1 AF results. Same and Cross average five and two evaluations; Weighted all uses $5:2$ weighting. Lower is better.}
\label{tab:af}
\scriptsize
\setlength{\tabcolsep}{3.5pt}
\begin{tabular}{l|c|c|c}
\hline
Method & Same (5 evals.) & Cross (2 evals.) & Weighted all \\
& AF(mAP/R1) $\downarrow$ & AF(mAP/R1) $\downarrow$ & AF(mAP/R1) $\downarrow$ \\
\hline
AKA (CVPR 2021)~\cite{pu2021aka} & 12.9/13.8 & 5.0/6.8 & 10.6/11.8 \\
LSTKC (AAAI 2024)~\cite{xu2024lstkc} & 15.9/13.0 & 7.5/9.3 & 13.5/11.9 \\
USP (TCSVT 2024)~\cite{yan2024unified} & 13.6/11.8 & 5.6/7.3 & 11.3/10.5 \\
PatchKD (ACM MM 2022)~\cite{sun2022patchkd} & 6.3/6.3 & 4.5/5.4 & 5.8/6.0 \\
DKP (CVPR 2024)~\cite{xu2024dkp} & 7.8/6.2 & 3.0/4.8 & 6.4/5.8 \\
DKC-controlled & \underline{3.3}/\underline{2.4} & \textbf{0.0}/\textbf{0.0} & \underline{2.4}/\underline{1.7} \\
DADKC (Ours) & \textbf{2.4}/\textbf{1.9} & \underline{0.2}/\underline{0.3} & \textbf{1.8}/\textbf{1.4} \\
\hline
\end{tabular}
\end{table}

On Order-1, weighted AF decreases from 2.4/1.7 to 1.8/1.4. On Order-2, aggregate AF changes from 2.7/2.1 to 2.0/1.5, and BWT improves from \mbox{$-2.3$/$-1.8$} to \mbox{$-1.6$/$-1.2$}.

\subsection{Ablation and Parameter Analysis}

We conduct ablation studies on training \textit{Order-1} to verify the contribution of each DADKC component. Table~\ref{tab:analysis} reports single modules, pairwise combinations, feature distillation, a generic MoE, and a parameter-matched MLP. When APA is not used, PPM is implemented without anchor guidance and relies on the base feature and domain or unknown embedding for expert routing. Table~\ref{tab:hyper} further analyzes the influence of DADKC-specific hyper-parameters, including the anchor pool size $M$, the number of retrieved anchors $K$, the number of experts per stage, and the loss weights $\lambda_{apa}$, $\lambda_{ppm}$, and $\lambda_{sba}$.

\begin{table}[!t]
    \centering
    \caption{Order-1 component and capacity analysis. DKC-reported is the published reference; other rows are common-pipeline five-seed variants.}
    \label{tab:analysis}
    \scriptsize
    \setlength{\tabcolsep}{4.5pt}
    \renewcommand{\arraystretch}{1.08}
    \begin{tabular}{l|l|cc}
        \hline
        Variant & Purpose & Avg. mAP & Avg. R1 \\
        \hline
        DKC-reported & published reference & 44.2 & 55.8 \\
        DKC + matched MLP & capacity control & 44.5 & 56.0 \\
        DKC + generic feature KD & alignment control & 44.6 & 56.3 \\
        DKC + generic MoE & routing control & 44.8 & 56.4 \\
        DKC + APA & anchor only & 44.7 & 56.3 \\
        DKC + PPM & path only & 44.9 & 56.5 \\
        DKC + SBA & bridge only & 44.6 & 56.4 \\
        DKC + APA + PPM & no SBA & \underline{45.3} & 57.0 \\
        DKC + APA + SBA & no PPM & 45.1 & 56.9 \\
        DKC + PPM + SBA & no APA & 45.2 & \underline{57.1} \\
        Full DADKC & all modules & \textbf{45.7} & \textbf{57.6} \\
        \hline
    \end{tabular}
\end{table}

\begin{table}[!htbp]
    \centering
    \caption{Order-1 sensitivity. $^\dagger$ marks the default used by every main table.}
    \label{tab:hyper}
    \scriptsize
    \setlength{\tabcolsep}{4.5pt}
    \renewcommand{\arraystretch}{1.08}
    \begin{tabular}{l|l|cc}
        \hline
        Parameter & Values & Avg. mAP & Avg. R1 \\
        \hline
        $M$ & $64 / 128^\dagger / 256$ & 45.3 / \textbf{45.7} / \underline{45.4} & 57.1 / \textbf{57.6} / \underline{57.2} \\
        $K$ & $4 / 8^\dagger / 12$ & 45.1 / \textbf{45.7} / \underline{45.5} & 56.9 / \textbf{57.6} / \underline{57.3} \\
        $E$ & $3 / 5^\dagger / 7$ & \underline{45.4} / \textbf{45.7} / 45.3 & \underline{57.2} / \textbf{57.6} / 57.1 \\
        $\lambda_{apa}=\lambda_{ppm}$ & $.05 / .10^\dagger / .20$ & 45.3 / \textbf{45.7} / \underline{45.4} & 57.2 / \textbf{57.6} / \underline{57.3} \\
        $\lambda_{sba}$ & $.005 / .01^\dagger / .02$ & 45.4 / \textbf{45.7} / \underline{45.5} & 57.3 / \textbf{57.6} / \underline{57.4} \\
        \hline
    \end{tabular}
\end{table}

\begin{table}[t]
    \centering
    \caption{Resource profile after five stages. Training memory includes the frozen SBA teacher; latency is single-image inference after warm-up.}
    \label{tab:efficiency}
    \scriptsize
    \setlength{\tabcolsep}{4.5pt}
    \renewcommand{\arraystretch}{1.08}
    \begin{tabular}{l|ccccc}
        \hline
        Method & Params & GFLOPs & Train GB & ms/image & Growth/stage \\
        \hline
        DKC & 28.7M & 4.11 & 7.2 & 4.8 & 0 \\
        Matched MLP & 31.1M & 4.45 & 7.8 & 5.2 & 0 \\
        DADKC & 31.2M & 4.48 & 9.9 & 5.5 & 0.48M \\
        \hline
    \end{tabular}
\end{table}

\subsection{Efficiency and Scalability}

Compared with DKC, the final model adds 8.7\% parameters, 9.0\% GFLOPs, and 14.6\% latency. The expert bank grows linearly with the number of stages, so longer streams can merge experts or use a fixed-capacity budget when strict storage is required. Table~\ref{tab:efficiency} summarizes the resource profile.

These results indicate that DADKC's improvement is not mainly obtained by heavy capacity expansion. The parameter-matched MLP uses similar capacity, but only reaches 44.5\%/56.0\% mAP/R1 compared with DADKC's 45.7\%/57.6\% mAP/R1 in Table~\ref{tab:analysis}, showing the benefit of APA, PPM, and SBA. In particular, the small increase from 31.1M to 31.2M parameters brings a larger accuracy gain than the MLP capacity control, which supports the effectiveness of structured anchor retrieval and stage-aware routing. The main extra cost is training memory from the frozen SBA teacher, while inference remains 5.5 ms/image. Thus, DADKC keeps the deployment cost close to DKC and pays most of its overhead only during continual-stage optimization.

\subsection{Visualization}
\begin{figure}[t]
    \centering
    \includegraphics[width=1.0\textwidth]{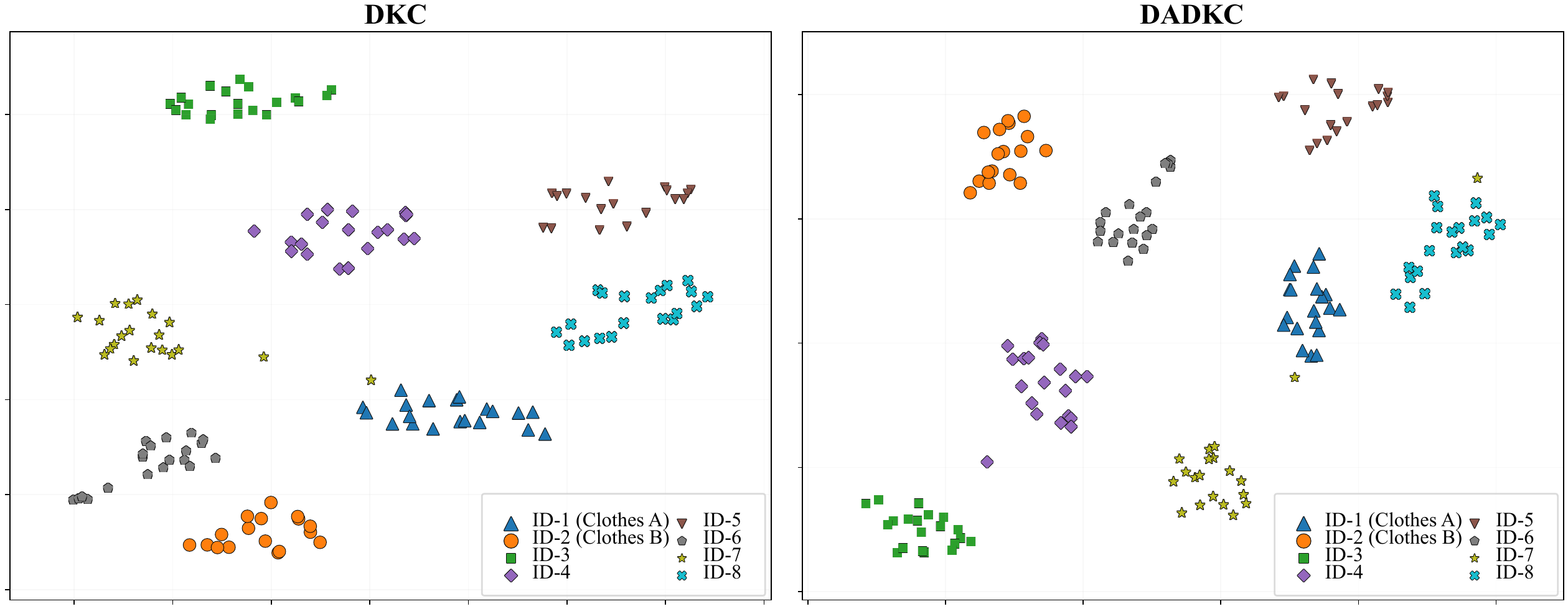}
    \caption{t-SNE visualization of learned representations on training \textit{Order-1}. Different colors represent different identities, and $\triangle$, $\bigcirc$ denote two randomly sampled identities with different clothes.}
    \label{fig:tsne}
\end{figure}

\begin{figure}[t]
    \centering
    \includegraphics[width=1.0\textwidth]{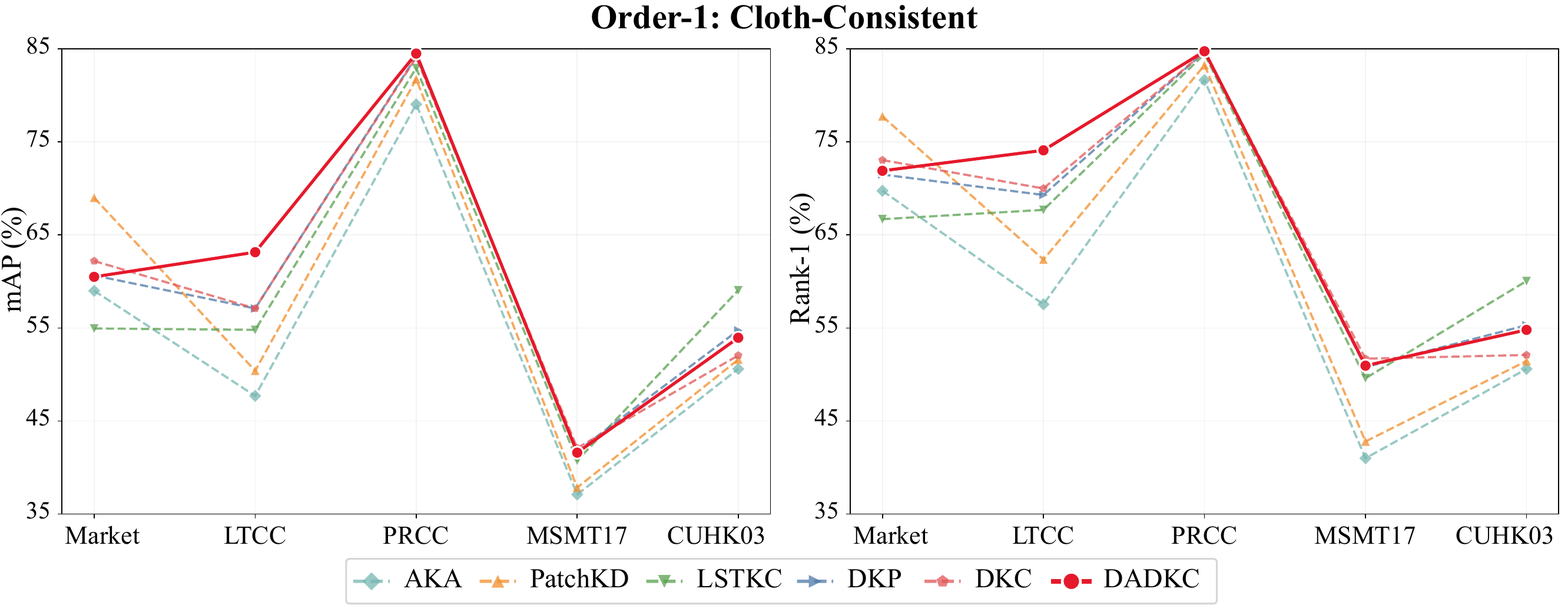}
    \caption{Performance tendency at different training stages on \textit{Order-1}.}
    \label{fig:tendency}
\end{figure}

To intuitively assess DADKC's representation quality, we visualize feature distributions via t-SNE at different training stages (Fig.~\ref{fig:tsne}), comparing intra-class compactness and inter-class separability under the cloth-consistent setting as in DKC \cite{cui2025dkc}. We also plot stage-wise performance trends in Fig.~\ref{fig:tendency} to examine whether DADKC steadily preserves historical knowledge as stages increase.

Together, Figs.~\ref{fig:tsne} and~\ref{fig:tendency} distinguish embedding geometry from retention across stages, where domain shifts and clothing changes can reshape the feature space. This complements Tables~\ref{tab:af} and~\ref{tab:analysis}: lower forgetting indicates better historical retention, while removing APA, PPM, or SBA weakens aggregate performance.

Fig.~\ref{fig:tendency} shows that DKC is competitive in early stages, but DADKC achieves higher mAP and Rank-1 accuracy in all later phases. This sustained advantage supports APA and PPM's ability to balance domain-aware knowledge without severely eroding historical information. In Fig.~\ref{fig:tsne}, DADKC produces tighter identity clusters and clearer inter-class margins than DKC, indicating better feature organization beyond classifier changes. With SBA, it also shows less cross-stage representation drift, supporting robust lifelong knowledge consolidation.

\section{Conclusion}


In this paper, we explore a practical and challenging problem called Cloth-Hybrid Lifelong Person Re-identification (CH-LReID), which requires retrieving the same person based on sequentially collected cloth-consistent and cloth-changing data while handling domain shifts and clothing variations. To tackle the above issue, we propose a Distribution-Aware Adaptive Differentiated Knowledge Consolidation (DADKC) framework. Building upon the DKC paradigm, DADKC introduces three lightweight modules: an Anchor Pool Adapter (APA) to refine identity features, a Phase Path Mixer (PPM) to integrate stage-aware domain distributions via expert routing, and a Stage Bridge Alignment (SBA) module to regularize cross-stage representation continuity. 
These modules enhance distribution modeling and cross-stage feature consistency without storing historical images or requiring domain labels at inference.
Experiments under the evaluated CH-LReID and traditional LReID protocols show higher average retrieval accuracy than the DKC baseline and lower aggregate forgetting on the tested CH-LReID streams.

\begin{credits}
\subsubsection{\discintname}
The authors declare that they have no competing interests.
\end{credits}

\clearpage
\bibliographystyle{splncs04}
\bibliography{dadkc_references}
\end{document}